\PassOptionsToPackage{numbers,sort&compress}{natbib}
\documentclass{WileyMSP-template}

\usepackage{amsmath}
\usepackage{amssymb}
\usepackage{graphicx}
\usepackage{xcolor}
\usepackage[font=sf,labelfont=bf]{caption}
\usepackage{natbib}
\usepackage{hyperref}
\usepackage{float}

\begin{document}

\pagestyle{fancy}
\rhead{\includegraphics[width=2.5cm]{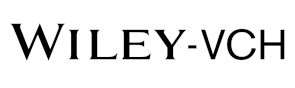}}

\title{Shake to Learn: Dynamic Interrogation of Hidden Object Physics for Robotic Manipulation with Physical Reservoir Computing}

\maketitle

\author{Wen Sin Lor$^\dagger$*}
\author{Jun Wang$^\dagger$}
\author{Suyi Li*}

\begin{affiliations}
W. S. Lor, Prof. S. Li\\
Department of Mechanical Engineering, \\ Virginia Tech, Blacksburg, VA 24060, USA\\[6pt]

Dr. J. Wang\\
Department of Mechanical Engineering, \\University of Michigan, Ann Arbor, MI 48109, USA\\[6pt]
$^\dagger$W. S. Lor and J. Wang contributed equally to this work.\\[6pt]

*Correspondent email: \texttt{wensin@vt.edu, suyili@vt.edu}
\end{affiliations}


\begin{abstract}
Many physical properties relevant to robotic manipulation are hidden from vision. A sealed object, for example, may reveal little about its center of mass (COM) or internal contents until it is lifted, shaken, or otherwise dynamically perturbed. This study shows that such interactions can enable a new modality of robotic perception and learning, in which interaction-induced dynamic responses are used to infer object physics that is inaccessible to conventional sensing. We implement this idea using an origami-inspired soft robotic arm that functions as a physical reservoir computer. After grasping an object, the arm is excited by a fixed shaking input at its base, and the resulting ringdown response is recorded through either camera tracking or embedded sensors. Because the input is held constant across trials, hidden object properties, such as the COM position, are encoded through their effect on the dynamics of the coupled robot-object system. A lightweight linear readout can then decode these dynamics to recover interpretable information about the hidden object physics. Using this framework, the soft robotic arm reservoir completed three tasks of increasing difficulty: inferring the orientation of the object’s hidden COM, inferring the COM distance from the grasp point, and using the inferred COM information to guide a subsequent regrasp. We further develop a dynamic summary representation of the ringdown response that improves prediction accuracy. Together, these results establish shake-to-learn mechanical interrogation as a promising strategy for robotic systems to convert brief physical interactions into actionable cues about hidden object properties for downstream manipulation.
\end{abstract}

\section{Introduction}
When manipulating a complex object, the robot often needs to reason about its physical characteristics not directly accessible through conventional perception modalities such as vision or haptics. For example, while the object’s shape and pose reveal its geometry and location, they provide little information about how its mass is distributed, how its internal contents are arranged, or how it will generate reaction forces and torques during manipulation. A sealed delivery box may appear uniform externally even though its contents are concentrated entirely on one side, and a transparent plastic bottle provides few visual cues about whether the liquid inside is highly viscous or free-flowing. These latent physical properties can be crucial for the success of robotic manipulation, determining whether a grasp remains stable or fails by tipping and slipping~\cite{kanoulas2018center}. Yet they are inherently difficult to infer from static appearance or surface contact alone. This limitation motivates the central question of this work: \textit{How can robots make hidden physical state observable, measurable, and interpretable}?

\medskip
Humans addressed this challenge by integrating two capabilities: dynamic physical interaction and distributed sensorimotor information processing. Rather than relying solely on visual observation, we lift, shake, tilt, and feel objects to solicit dynamic responses that reveal information unavailable from appearance alone~\cite{lederman1987hand}. These interaction-generated dynamic cues encode a wealth of latent physical properties. For example, an offset of center of mass (COM) is perceived as a torque during lifting~\cite{schneider2022object, rens2021lift}; loose internal contents produce delayed or oscillatory motion during shaking~\cite{sekiguchi2003haptic, tanaka2012shaking, frissen2023humans}; and compliant or fluid-filled objects reveal themselves through deformation and flow dynamics under manipulation~\cite{bergmann2009cues, hauser2018force, jansson2006liquid}. Equally important, the brain isn't the only structure interpreting these dynamic cues. Human manipulation relies on hierarchical and distributed sensorimotor processes---spanning from body mechanics, tactile afferents, and spinal reflex pathways to cortical circuits---that enable contact-induced dynamics to support rapid perception, continuous state estimation, and timely motor adaptation~\cite{prochazka2000reflex, loeb1999hierarchical, johansson2009coding}. This biological strategy suggests that the body itself also plays a central role in learning and perception by actively \textit{projecting} hidden object properties into feature-rich dynamic cues and then efficiently \textit{extracting} this information for reasoning and task planning.

\medskip
Translating this human strategy to robotics suggests a new learning modality comprising two tightly coupled processes: encoding (projection) and decoding (extraction). During encoding, active physical interactions project an object’s latent physical properties onto \textit{measurable} dynamic responses. During decoding, these responses are processed to extract interpretable information that subsequently informs robotic tasks. This framework presents two primary challenges. First, encoding must be sufficiently feature-rich and sensitive to reveal complex hidden physics. Second, decoding must be fast and efficient, extracting relevant information with minimal latency to support dynamic manipulation. However, prior approaches to interaction-based robotic learning have struggled to meet these requirements simultaneously. Their encoding mechanisms are often low-dimensional (e.g., using simple contact-force measurements~\cite{kanoulas2018center,wang2021parameter}), limiting their ability to capture complex physical properties; meanwhile, their decoding pipelines rely heavily on digital computation (e.g., separately trained machine-learning models~\cite{wang2020swingbot,ai2024robopack}), introducing substantial overhead and latency.

\medskip
In this study, we aim to address these two challenges simultaneously by exploiting the framework of physical reservoir computing (PRC), \emph{which enables us to implement rich encoding and efficient decoding into a unified physical architecture}. In PRC, a physical substrate serves as a nonlinear dynamical transformer that projects input signals into a high-dimensional state space, and a trained readout maps the resulting states to the desired outputs~\cite{jaeger2001echo, maass2002real, nakajima2021reservoir}. Consequently, the substrate’s deformation, nonlinearity, fading memory, and distributed dynamics become computational resources~\cite{tanaka2019recent, nakajima2020physical, stepney2024physical}. This computing principle has been demonstrated across a wide range of physical systems, including delayed dynamical nodes~\cite{appeltant2011information}, nonlinear oscillators~\cite{coulombe2017computing, perkins2025duffing}, distributed compliant system~\cite{hauser2011towards}, origami structure~\cite{wang2023building}, soft robotic bodies~\cite{nakajima2014exploiting, bhovad2021physical, wang2025proprioceptive}, emerging electronic devices~\cite{liang2024emerging}, active-particle media~\cite{wang2024active}, nanomechanical resonators~\cite{kartal2025nanomechanical}, and compliant fiber networks~\cite{khairnar2026spider}. Among these platforms, soft robotic reservoirs are particularly relevant because they demonstrate that compliant bodies can simultaneously serve as sensing interfaces and computational media through their own deformation and dynamics~\cite{wang2025proprioceptive, he2025softswimmer, terajima2025tensegrity, wang2026embodying}.

\begin{figure}[t]
  \centering
  \includegraphics[width=0.99\textwidth]{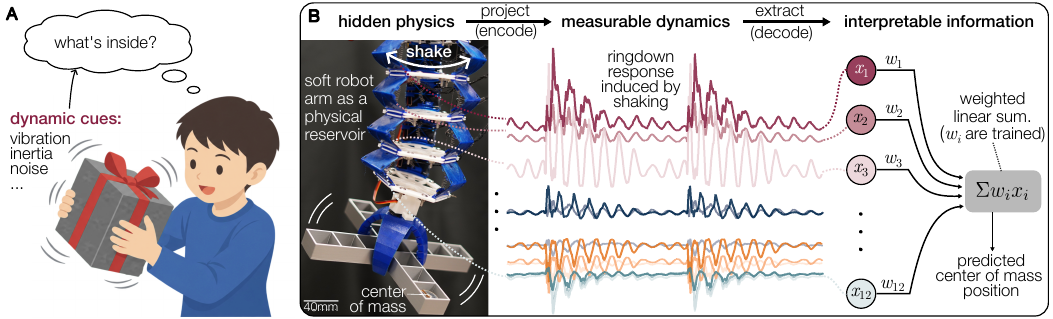}
  \vspace{-0.1in}
  \caption{\textbf{"Shake-to-learn" dynamic interrogation}---an overview of the proposed robotic learning modality based on interaction-induced dynamics. (\textbf{A}) The approach draws inspiration from how children physically manipulate an object to elicit dynamic cues and uses them to infer what's inside. (\textbf{B}) In the robotic analogue, a soft robotic arm---doubling as a physical reservoir computer---grasps an object whose physical state is visually hidden, delivers a brief dynamic probe, and records the resulting body response. The reservoir computing power embodied in the soft arm projects the robot–object interaction into a separable, high-dimensional ringdown response (i.e., reservoir states), from which a linear readout extracts the hidden physical properties. Here, we focus specifically on the object's center-of-mass (COM) position and how it shifts.}
  \label{fig-concept}
\end{figure}

\medskip
Therefore, in this work, we demonstrate the new robotic learning modality via dynamic interaction---feature-rich encoding and efficient decoding of latent object properties---by repurposing a modular, origami-inspired robotic arm as a physical reservoir computer~\cite{wang2025repurposing}. Specifically, we implement this strategy through a ``shake-to-learn'' dynamic interrogation protocol, inspired by how small children like to shake their Christmas boxes first to guess what their gifts are (Figure~\ref{fig-concept}). After grasping an object, the robotic arm receives a fixed, servo-driven impulse at its base, and the resulting ringdown dynamics are measured using either an external camera or embedded bending sensors.

\medskip
Although the robotic arm remains mechanically unchanged, grasping an object fundamentally alters the dynamics of the coupled robot–object system. The object modifies the effective inertia, gravitational loading, boundary conditions, equilibrium configuration, and vibration characteristics. Consequently, the same mechanical impulse serves as a repeatable interrogation probe: the excitation is held constant, while different latent object states produce distinct ringdown responses. These responses form a dynamical signature of the object’s hidden physical state. By training a linear readout on these signatures (i.e., without a separately trained digital learning model), one can accurately estimate the object’s hidden center-of-mass (COM) position, thus providing information for downstream manipulation tasks including re-grasping, balancing, and motion planning.

\medskip
More broadly, this work points toward a new design principle for intelligent robots: compliant structures can be engineered not only to actuate and sense, but also to compute and learn. By leveraging dynamic interaction and physical reservoir computing, these embodied systems can transform hidden physics into informative dynamical signals that are directly interpretable for robotic perception. This new robotic learning modality could complement current paradigms based on vision, haptics, and foundational models, ultimately helping to bridge the ``physics gap'' that limits robotic perception and adaptability \cite{alu2025roadmap}.

\section{Dynamic Interrogation: Task Setup and Theoretical Framework}
\label{sec:mechanical_interrogation_framework}
\begin{figure}[t!]
  \centering
  \includegraphics[scale=1.0]{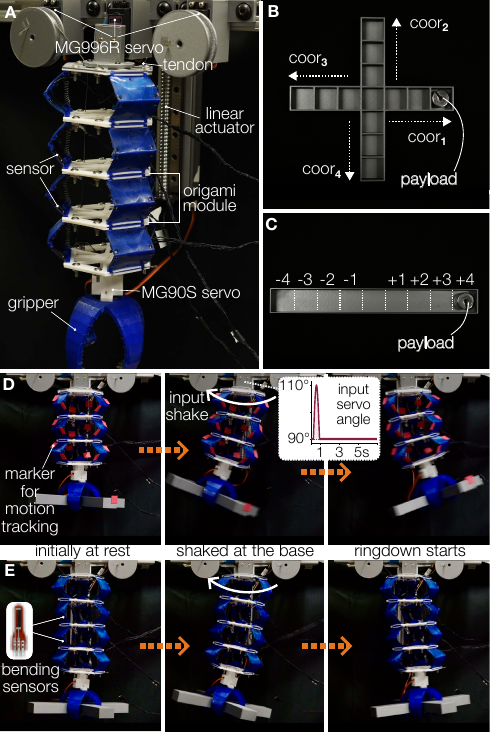}
  \caption{\textbf{Experiment setup for dynamic interrogation}.
  (\textbf{A}) Anatomy of the modular soft robotic arm/physical reservoir.
  (\textbf{B}) 3D-printed cross-shaped container, showing the hidden payload mass $m$ and the corresponding coordinate setup. 
  (\textbf{C}) 3D-printed bar-shaped container. 
  (\textbf{D}) The ringdown test procedure on a three-module robotic arm using camera-based motion tracking.
  (\textbf{E}) Another test on a four-module arm using embedded bending strain sensors. 
  }
  \label{fig-setup}
\end{figure}

Dynamic interrogation aims to project hidden object properties into measurable dynamical responses in the coupled robot-object system. In this study, we choose the object's COM as the targeted latent property. To this end, we 3D-print a cross-shaped and a bar-shaped container with internal dividers, and then place a payload mass $m$ inside at a selected location $\ell$ (Figure \ref{fig-setup}). Therefore, adjusting the payload's location inside the container effectively changes the object's COM.

\medskip
Once the soft robotic arm grasps the container, a brief servo-driven impulsive actuation (i.e., shake) is applied at the arm’s base. The actuation ceases so that the coupled system undergoes a passive ringdown. Because the impulse excitation is the same across different trials, any variations in the ringdown responses directly reflect how the hidden COM modifies the system dynamics. 

\medskip
In what follows, we explain the experimental apparatus, underlying theory, and physical reservoir training for setting up this dynamic interrogation task. 

\subsection{Experiment Setup: Doubling a Soft Robotic Arm as a Physical Reservoir}
\label{subsec:experimental_setup_fixed_protocol}
The experimental platform consists of a modular origami robotic arm mounted on an aluminum frame and equipped with a soft gripper at its distal end (Figure \ref{fig-setup}A). Four servomotors actuate the robotic arm. A top-mounted MG996R servomotor generates the mechanical impulse that shakes the coupled robotic arm and object. Two side-mounted MG996R servomotors, each connected to a Kevlar tendon, bend the robotic arm during the grasping and regrasping sequence. Finally, an MG90S servomotor actuates the soft gripper. These servomotors execute pre-defined, open-loop motion primitives. That is, the physical reservoir does not learn low-level motor commands; instead, it infers hidden object properties that determine which motion primitive to execute.

\medskip
To measure the ringdown dynamics (i.e., reservoir states), we employed two sensing strategies. The first strategy relied on high-resolution vision. A three-module robotic arm was instrumented with 19 sparsely distributed visual markers (Figure \ref{fig-setup}D). During experiments, the arm dynamics were recorded using a Sony $\alpha$7C camera operating at 30 Hz, and an image-processing pipeline tracked the markers' positions, yielding 38 two-dimensional marker coordinate features per frame.

\medskip
The second strategy employed embedded bending sensors, enabling fully onboard, real-time operation (Figure \ref{fig-setup}E). Twelve bi-directional bending strain gauges were integrated along a four-module robotic arm, and their outputs were acquired using an NI cDAQ-9178 chassis equipped with an NI 9205 voltage-input module. Additional details of hardware and data acquisition setup are available in SI Section S1.

\subsection{Theoretical Foundation: Why COM Position is Encoded into Reservoir Dynamics}
\label{subsec:contact_parametric_ringdown_mechanism}
The key distinction from our proposed dynamic-interrogation framework and more conventional information processing with physical reservoir computing is how the targeted information enters the dynamic responses. A physical reservoir can be written schematically as

\begin{equation}
    \dot{\mathbf{x}}(t)
    =
    F\!\left(\mathbf{x}(t),\mathbf{u}(t);\boldsymbol{\theta}\right),
\end{equation}

where \(\mathbf{x}(t)\) is the measured reservoir state, \(\mathbf{u}(t)\) is the external excitation, and \(\boldsymbol{\theta}\) denotes the physical parameters of the reservoir.  

\medskip
In the conventional setting, the input information enters the system via customized and encoded excitation \(\mathbf{u}(t)\). In our framework, however, the external input is the same impulsive shake \(\mathbf{u}_{\mathrm{imp}}(t)\) for every trial. The object's latent physics instead enters the system by changing the physical parameters of the reservoir:

\begin{equation}
    \dot{\mathbf{x}}(t)
    =
    F\!\left(
    \mathbf{x}(t),
    \mathbf{u}_{\mathrm{imp}}(t);
    \boldsymbol{\theta}+\Delta\boldsymbol{\theta}(m,\ell)
    \right),
\end{equation}

where \(m\) is the object's mass and \(\ell\) is its position within the 3D-printed container, which then shifts the overall COM. Therefore, the object's latent physics is not supplied as a new input waveform, but changes how the reservoir rings down after the fixed shake.

\medskip
The physical origin of \(\Delta\boldsymbol{\theta}(m,\ell)\) follows the object-induced inertia perturbation and gravitational loading. Denote \(\mathbf{q}\) as the generalized coordinates of the robotic arm, and let \(\mathbf{r}_{\ell}(\mathbf{q})\) be the Cartesian position of the object's COM at location \(\ell\). With Jacobian

\begin{equation}
    \mathbf{J}_{\ell}(\mathbf{q})
    =
    \frac{\partial \mathbf{r}_{\ell}}{\partial \mathbf{q}},
\end{equation}

The object perturbs the mass matrix and gravitational potential of the coupled robot-object system in that

\begin{equation}
    \Delta \mathbf{M}(\mathbf{q},m,\ell)
    =
    m\mathbf{J}_{\ell}(\mathbf{q})^{\top}\mathbf{J}_{\ell}(\mathbf{q}),
    \qquad
    \Delta V(\mathbf{q},m,\ell)
    =
    mgh_{\ell}(\mathbf{q}).
    \label{eq-Delta}
\end{equation}

The first term changes the effective inertia seen by the arm, while the second changes the gravitational energy landscape. Both terms scale with payload mass $m$ and vary according to its location $\ell$, giving a direct physical basis for why heavier objects and more eccentric center-of-mass positions should produce more distinct ringdown responses.

\medskip
The camera tracking and bending sensor measurements are projections of the perturbed physical reservoir dynamics:

\begin{equation}
    \mathbf{x}_{\mathrm{cam}}(t)
    =
    \Phi_{\mathrm{cam}}(\mathbf{q}(t)),
    \qquad
    \mathbf{x}_{\mathrm{bend}}(t)
    =
    \Phi_{\mathrm{bend}}(\mathbf{q}(t)).
\end{equation}

Camera tracking observes the distributed robotic arm motions, while the bending sensors observe local material deformations. If both sensing routes reveal the hidden mass center's position, we can conclude that the latent properties are encoded in the arm--object ringdown dynamics, rather than in a sensor-specific artifact.

\medskip
Additionally, the measured response can be interpreted as a combination of load-induced equilibrium bias and oscillatory ringdown,

\begin{equation}
    \mathbf{x}_s(t;m,\ell)
    =
    \mathbf{x}_{\mathrm{eq},s}(m,\ell)
    +
    \delta\mathbf{x}_{\mathrm{osc},s}(t;m,\ell),
    \qquad
    s\in\{\mathrm{cam},\mathrm{bend}\}.
    \label{eq-ringdown}
\end{equation}

Here, \(\mathbf{x}_{\mathrm{eq},s}(m,\ell)\) represents the load-induced equilibrium deformation, while \(\delta\mathbf{x}_{\mathrm{osc},s}(t;m,\ell)\) represents the oscillatory ringdown. Because both terms depend on \(m\) and \(\ell\), changes in payload mass and location alter the response observed through either sensing strategy. The complete derivation of this ringdown model is provided in SI Section~S2.

\subsection{Weighted Linear Readout: How Center of Mass (COM) Position Is Decoded}
\label{subsec:task_encoding_linear_decoding}
We keep the reservoir readout layer deliberately simple so that dynamic interrogation remains tied to the physical responses of the robotic arm and object. Specifically, each ringdown experiment trial produces a temporal measurement of the physical reservoir's internal states:

\begin{equation}
    \mathbf{X}^{(i)}
    =
    \left[
    \mathbf{x}^{(i)}_1,
    \mathbf{x}^{(i)}_2,
    \ldots,
    \mathbf{x}^{(i)}_T
    \right]^{\top},
    \label{eq-reservoir-states}
\end{equation}
 
where the superscript \((i)\) indexes the trial, and \(T\) is the number of sampled time frames within the observation window. \(\mathbf{x}^{(i)}_t \, (t=1\ldots T)\) is the measured reservoir state vector at time frame \(t\), which can be either a camera-tracked marker displacement vector or a voltage vector from the embedded bending sensors. To decode
these measured responses, the linear-readout weights are fitted by ordinary least squares using Eq.~\eqref{eq-readout-training}. 

\begin{equation}
    \left(
        \mathbf{W}_{\mathrm{out}}^{*}
    \right)^{\top}
    =
    \mathbf{Y}_{\mathrm{train}}
    \begin{bmatrix}
        \mathbf{1}_{T}^{\top} & \cdots & \mathbf{1}_{T}^{\top}\\
        \left(\widetilde{\mathbf{X}}^{(1)}\right)^{\top}
        & \cdots &
        \left(\widetilde{\mathbf{X}}^{(N_{\mathrm{tr}})}\right)^{\top}
    \end{bmatrix}^{\dagger}.
    \label{eq-readout-training}
\end{equation}

Here, \(\widetilde{(\cdot)}\) denotes a z-score normalized quantity, $N_{\mathrm{tr}}$ is the number of training trials, $\mathbf{Y}_{\mathrm{train}}$ contains the corresponding target vectors, $\mathbf{1}_{T}$ is a length-$T$ vector of ones representing the bias term, and $(\cdot)^{\dagger}$ denotes the Moore--Penrose pseudoinverse. Once trained, the readout weights are held fixed and applied to each frame of a held-out trial \(i\),

\begin{equation}
    \widehat{\mathbf{y}}_{t}^{(i)}
    =
    \left(
        \mathbf{W}_{\mathrm{out}}^{*}
    \right)^{\top}
    \begin{bmatrix}
        1\\
        \widetilde{\mathbf{x}}_{t}^{(i)}
    \end{bmatrix}.
    \label{eq-linear-readout}
\end{equation}

The frame-level outputs are averaged over the observation window to
obtain the output of each ringdown trial,

\begin{equation}
    \bar{\mathbf{y}}^{(i)}
    =
    \frac{1}{T}
    \sum_{t=1}^{T}
    \widehat{\mathbf{y}}_{t}^{(i)}.
\end{equation}

Here, the output $\bar{\mathbf{y}}^{(i)}$ is predicted COM offset. It is worth highlighting that the readout weights are trained with linear regression, without involving a separate digital learning model. The soft robotic arm itself, as the physical reservoir, serves as the computation kernel that projects the input shake into high-dimensional, nonlinear dynamical states. The trained linear readout extracts desired information from these states.

\medskip
Since this study uses two different 3D-printed containers as the interrogated object---cross-shaped and bar-shaped, we formulated the output \(\bar{\mathbf{y}}^{(i)}\) accordingly.

\medskip
For the cross-shaped container, the hidden payload mass $m$ is placed along one of the four angular directions, coor\(_1\)--coor\(_4\) (Figure \ref{fig-setup}B), and our goal is to use dynamic interrogation to infer the COM's \textit{angular orientation}. Correspondingly, we define four coordinates:

\begin{equation}
    \mathbf{c}_{1}=[1,0]^{\top},\quad
    \mathbf{c}_{2}=[0,1]^{\top},\quad
    \mathbf{c}_{3}=[-1,0]^{\top},\quad
    \mathbf{c}_{4}=[0,-1]^{\top}.
\end{equation}

Therefore, the output $\bar{\mathbf{y}}^{(i)}$ is defined with respect to these four coordinates. To classify the COM's orientation based on $\bar{\mathbf{y}}^{(i)}$, we can apply another layer of nearest-target decoding $\hat{k}^{(i)}=\arg\min\left\|\bar{\mathbf{y}}^{(i)}-\mathbf{c}_{k} \right\|_2$, where $k=1\ldots 4$.

\medskip
For the bar-shaped container (Figure~\ref{fig-setup}C), the hidden payload mass $m$ can be placed anywhere along its length, and our goal is to use dynamic interrogation to infer the COM's \textit{linear position}.  Therefore, we define an offset from the grasping location such that

\begin{equation}
    o=x_{\mathrm{COM}}-x_{\mathrm{grip}},
\end{equation}

where \(x_{\mathrm{COM}}\) is the object's COM (i.e., payload mass) position along the bar and \(x_{\mathrm{grip}}\) is the gripping position (Figure~\ref{fig-setup}). 
Correspondingly, we define a different coordinate 

\begin{equation}
    \mathbf{c}_{o}=[o,0]^{\top}.
\end{equation}

where $o$ can take values between -4 and 4. To classify the COM's position based on $\bar{\mathbf{y}}^{(i)}$, we can also apply nearest-target decoding $\hat{o}^{(i)}=\arg\min\left\|\bar{\mathbf{y}}^{(i)}-\mathbf{c}_{o} \right\|_2$. 

\section{Dynamic Interrogation: Results and Insights}
\label{sec:mechanically_guided_inference}
In this study, we task the soft robotic arm with inferring the hidden COM position through three challenges of increasing complexity. Task I asks whether shake-induced ringdown dynamics carry enough information to decode the angular position of a hidden payload mass within a cross-shaped container. This establishes the fundamental principles of robotic learning via dynamic interaction, and confirms that this learning modality is accessible through both external camera tracking and embedded sensing alone. Task II moves from angular classification to a more manipulation-relevant scenario, where the readout must infer COM shifts within a bar-shaped container by predicting its distance from the grasp point. Finally, Task III shows how information gathered through these dynamic interrogations can guide subsequent robotic action, enabling the soft robot to re-grasp the object and achieve a more balanced configuration.

\subsection{Task I: Interrogating the Orientation of COM}
\label{subsec:level1_payload_angle_decoding}
\begin{figure}[t!]
  \centering
  \includegraphics[width=0.98\linewidth]{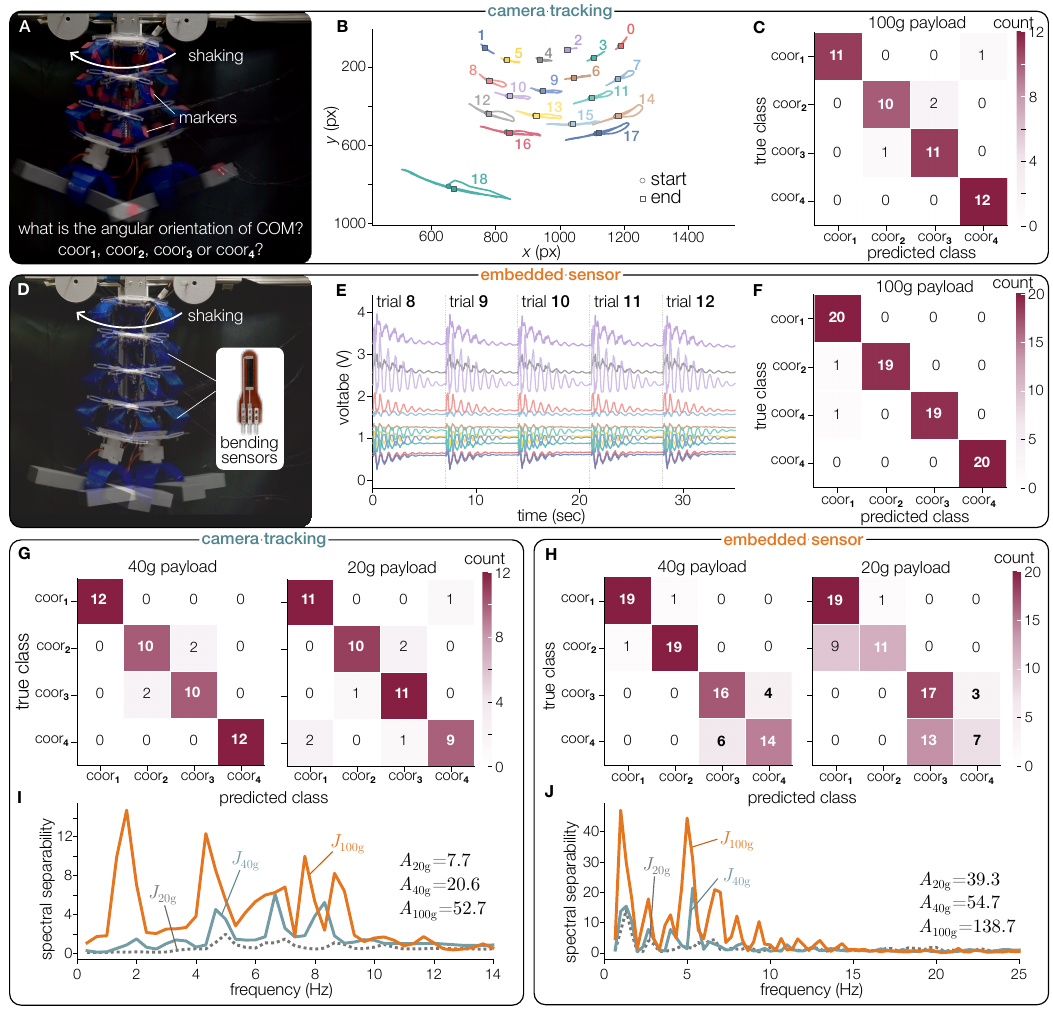}
  \caption{\textbf{Predicting the angular orientation of the hidden COM.} (\textbf{A--C}) Experimental snapshot, tracked marker displacement, and confusion matrix using external camera tracking. With a 100~g payload, the camera-tracking method yields 44/48 correct leave-one-out predictions. (\textbf{D--F}) Experimental snapshot, recorded sensor reading, and confusion matrix using embedded bending sensors. With a 100~g payload, the embedded bending sensors yield 78/80 correct leave-one-out predictions. (\textbf{G}) With 40~g and 20~g payloads, the camera-tracking strategy yields 44/48 and 41/48 correct predictions, respectively. (\textbf{H}) With 40~g and 20~g payloads, the embedded sensors yield 68/80 and 54/80 correct predictions, respectively. (\textbf{I, J}) Frequency-resolved spectral separability of the ringdown response across different COM orientations. A larger $J_m(f)$ indicates that the four COM-orientation classes produce more distinguishable frequency-domain ringdown signatures, while the area under the curve, $A_m$, summarizes the total separable spectral content across the frequency band. Trends in $J_m(f)$ and $A_m$ support the interpretation that lighter payloads produce weaker orientation-dependent spectral signatures in the shake-induced ringdown.
  }
  \label{fig-task1}
\end{figure}

This first task aims to validate that the hidden COM can indeed be encoded in the shake-induced ringdown dynamics and, importantly, that this capability is independent of the sensing strategy. To this end, Figure~\ref{fig-task1} summarizes and compares the results obtained using external camera-based motion tracking and embedded bending sensors.

\medskip
Note that camera-tracking experiments use a three-module arm instrumented with 19 motion markers, whereas the embedded-sensing experiments use a four-module robotic arm. The additional module is necessary because the bending sensors occupy substantially more physical space, limiting each module to only three sensors. Consequently, four modules, comprising 12 bending sensors in total, are used to provide a sufficiently high-dimensional reservoir state for encoding complex latent properties.

\medskip
In both cases, the robot receives the same brief input shaking the base, after which actuation ceases, and the resulting ringdown response is recorded. The readout then maps the measured ringdown dynamics to classify center-of-mass coordinates using the linear decoding rule defined in Section~\ref{subsec:task_encoding_linear_decoding}.

For every combination of payload mass $m$ and sensing strategy, we run multiple experimental trials and group these trials into four classes according to the angular coordinate of the true COM.  We used the balanced ``leave-one-trial-per-group'' (or LOO) validation scheme~\cite{kohavi1995study}. That is, one trial from each of the four classes was held out for testing, and the readout was trained on the remaining trials by utilizing Eq.~\eqref{eq-readout-training}. 

\medskip
This procedure was repeated until every trial in every coordinate class had been used once for testing. The reported accuracy counts all held-out predictions across these balanced folds. This training procedure is detailed in the SI Section S3A.

\medskip
With a $m=100$ gram hidden payload, the embedded bending sensing gives 78 correct leave-one-out predictions out of 80 trials, while camera tracking gives 44 correct predictions out of 48 trials (Figure \ref{fig-task1}C,F). Both sensing strategies, therefore, recover the hidden COM orientation from the same input shake. This agreement supports the interpretation that the hidden COM information is carried by the ringdown response of the coupled robot-object, and such information can be accessed through different sensing projections.

\medskip
Figure~\ref{fig-task1}(G,H) summarizes the effect of different payload mass on prediction performance. With embedded bending sensors, the number of correct predictions decreases from 78 out of 80 trials for the 100~gram payload to 68 out of 80 trials for the 40~gram payload and 54 out of 80 trials for the 20~gram payload. A similar, although less pronounced, trend is observed with camera-marker tracking, for which the corresponding numbers of correct predictions are 44, 44, and 41 out of 48 trials, respectively. This degradation in prediction performance with decreasing payload mass is expected, as lighter payloads exert weaker inertial and gravitational perturbations on the soft robotic arm, resulting in less separable ringdown signatures (Eq. \ref{eq-Delta}).

\medskip
To quantify this separability directly in the ringdown measurement, we computed the frequency-resolved spectral Fisher ratio of the ringdown response from with each payload mass, as shown in Figure~\ref{fig-task1}(I,J). For trial \((i)\) with payload mass \(m\), the ringdown response is converted to a multichannel FFT magnitude vector,

\begin{equation}
    \mathbf{s}^{(i,m)}(f)
    =
    \log\left(
        1+
        \left|
            \operatorname{FFT}
            \left\{\mathbf{x}_{t}^{(i,m)}\right\}(f)
        \right|
    \right).
\end{equation}

where \(\mathbf{s}^{(i,m)}(f)\in\mathbb{R}^{c}\) contains the spectral magnitude across \(c\) sensing channels at frequency \(f\). For each payload-orientation class \(k\in\{1,2,3,4\}\), the class-mean spectrum is

\begin{equation}
    \boldsymbol{\mu}^{(k,m)}(f)
    =
    \frac{1}{N_k}
    \sum_{i:y_i=k}
    \mathbf{s}^{(i,m)}(f),
\end{equation}

and the global mean spectrum is \(\boldsymbol{\mu}^{(m)}(f)\). We define the frequency-resolved COM-orientation separability as

\begin{equation}
    J_m(f)
    =
    \frac{
    \frac{1}{N}
    \sum_{i=1}^{N}
    \frac{1}{c}
    \left\|
    \boldsymbol{\mu}^{(y_i,m)}(f)
    -
    \boldsymbol{\mu}^{(m)}(f)
    \right\|_2^2
    }{
    \frac{1}{N}
    \sum_{i=1}^{N}
    \frac{1}{c}
    \left\|
    \mathbf{s}^{(i,m)}(f)
    -
    \boldsymbol{\mu}^{(y_i,m)}(f)
    \right\|_2^2
    +
    \epsilon
    }.
\end{equation}

This ratio compares the spectral difference among COM-orientation classes, while considering the trial-to-trial variability within the same class. A larger value therefore indicates that the COM-orientation produces a more reliable frequency-domain signature at that frequency. The integrated separability,

\begin{equation}
    A_m
    =
    \int_{f_1}^{f_2}
    J_m(f)\,df,
\end{equation}

quantifies the total separable spectral content between different COM orientations over the analyzed frequency band.

\medskip
Together, prediction results and frequency-resolved separability analysis in Figure~\ref{fig-task1} establish the basis for using dynamic interrogation to infer a hidden COM: A servo-driven input shake encodes the angular orientation of the payload mass into the coupled robot-object's ringdown response, and this information can be decoded using either camera-based marker tracking or embedded bending sensors.

\medskip
Notably, although the payload mass is placed at the four tips of the cross-shaped container in the experiments, its orientation remains decodable when it is moved closer to the container’s center. Direct transfer from the outer to the near radius is less accurate, while limited calibration at the near radius substantially improves prediction accuracy (SI Sections S4B and S4C). Moreover, the ability to infer hidden COM orientation is not limited to the soft configuration used in the main experiments. The hidden COM orientation can also be decoded when the robotic arm operates in stiff and mixed bistable configurations (SI Section S4A). Building on these results, we next examine how dynamic interrogation can localize a hidden COM with a higher resolution.

\subsection{Task II: Interrogating the Position and Distance of Hidden COM}
\label{subsec:level2_distance_aware_com}
\begin{figure}[t!]
    \centering
    \includegraphics[width=0.98\linewidth]{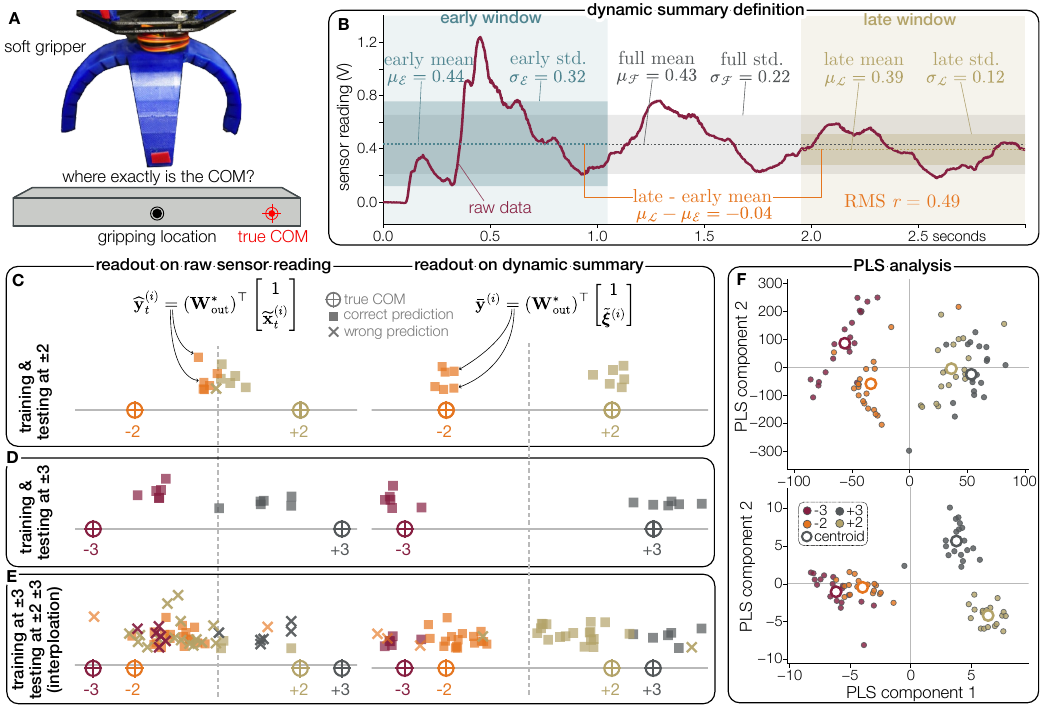}
    \caption{
    \textbf{Predicting the offset of hidden COM relative to the grasping point.}
     (\textbf{A}) A robotic grasp centered on an object can perform poorly if the true COM is located far away---a common challenge in realistic robotic manipulation tasks.
    (\textbf{B}) Definition of the dynamic summary, illustrated on a raw ringdown response from an embedded sensor. The early and late windows are shaded with different colors.
    (\textbf{C}) The physical reservoir's prediction of COM position using raw sensor data (left) and the dynamic summary (right). The prediction performance is also assessed using the "Leave-One-Out" (LOO) protocol: we conducted 12 experimental trials, all with true center-of-mass (COM) positions at $+2$ or $-2$. We held out one trial for testing and used the remaining trials for readout training; the outcome $\bar{\mathbf{y}}^{(i)}$ for this held-out trial is shown as a marker in the plot. To classify whether this $\bar{\mathbf{y}}^{(i)}$ prediction is correct, we also apply nearest-target decoding $\hat{o}^{(i)}=\arg\min\left\|\bar{\mathbf{y}}^{(i)}-\mathbf{c}_{o} \right\|_2$, where $o=\pm2, \pm3$ in this Task II.
    We repeated this process until every trial had been used for testing once.
    (\textbf{D}) Analogous test results where the true COM is at $+3$ or $-3$.
    (\textbf{E}) An additional set of test results illustrating interpolation. Here, the readout is trained using trials with COM at $\pm3$ only, but the testing trials can have COM at $\pm2$ or $\pm3$.
    (\textbf{F}) PLS analysis of raw sensor data (up) and dynamic-summary features (bottom).  
    }
    \label{fig-task2}
\end{figure}

Task II shifts the objective from decoding the hidden COM's angular orientation to accurately classifying its distance from the grasp location. In this task, the interrogated object is the bar-shaped container. Its uniform external geometry suggests a grasp at its geometric center; however, its hidden mass distribution may have shifted the true COM away from this location (Figure~\ref{fig-task2}A). Therefore, effective dynamic interrogation must determine not only which side of the bar-shaped container is heavier, but also how far the grasp point should adjust to achieve a more balanced outcome. This task setup captures a common challenge in robotic manipulation: a grasp that appears appropriate based solely on an object’s external, visible geometry may become unstable once its underlying mass distribution is considered. In addition, we will use embedded sensor readings hereafter to represent a more realistic robotic manipulation setup.

\medskip
First, we apply the readout directly to the raw bending sensor data, as in Task~I. These raw data preserve the frame-by-frame ringdown dynamics following the input shake. Positive and negative COM shifts tend to produce distinct deformation patterns, since they load the robotic arm in opposite directions. As shown in the left column of Figure~\ref{fig-task2}(C, D), this raw representation reliably captures the COM's \textit{orientation}---consistent with the conclusion from Task~I---but struggles to accurately predict its \textit{position}, yielding widely dispersed results that often deviate substantially from the true COM location. The raw sensor readings therefore contain physical information about the hidden COM, but this information is not structured in a form that a linear readout can reliably use to extract the COM position.

\medskip
To address this issue, we introduce a physically interpretable \textit{dynamic-summary} representation for accurately decoding the COM position. Rather than applying the readout to sensor readings from individual time frames, this representation describes each ringdown using statistics computed over selected portions of the response. Using summary statistics instead of raw temporal data has precedent in time-series machine learning. For example, the Time Series Forest method characterizes sampled intervals by their mean, standard deviation, and slope~\cite{deng2013timeseriesforest}. Here, we extend this idea to the hidden-COM interrogation by computing these statistics over the complete ringdown of the selected 3-second window, the first 35\% of the ringdown, and the final 35\%, respectively. These statistical signatures form a mechanics-informed, trial-level representation that compresses each ringdown into a single compact vector for subsequent linear readout. More specifically, the dynamic-summary vector for trial \((i)\) is defined as

\begin{equation}
    \boldsymbol{\xi}^{(i)}
    =
    \mathrm{concat}_{c}
    \left[
    \mu_{c,\mathcal{F}}^{(i)},
    \sigma_{c,\mathcal{F}}^{(i)},
    \mu_{c,\mathcal{E}}^{(i)},
    \sigma_{c,\mathcal{E}}^{(i)},
    \mu_{c,\mathcal{L}}^{(i)},
    \sigma_{c,\mathcal{L}}^{(i)},
    \mu_{c,\mathcal{L}}^{(i)}-\mu_{c,\mathcal{E}}^{(i)},
    r_c^{(i)}
    \right]^{\top},
    \label{eq-summary}
\end{equation}

where \((i)\) indexes the experimental trial and \(c\) indexes the bending-sensor channel. The windows \(\mathcal{F}\), \(\mathcal{E}\), and \(\mathcal{L}\) denote the full, early, and late ringdown response, respectively. The quantities \(\mu\) and \(\sigma\) are the mean and standard deviation computed over the corresponding window, while \(r_c^{(i)}\) is the RMS magnitude of channel \(c\) over the selected 3-second window. Based on this setup, the predicted COM offset becomes

\begin{equation}
    \bar{\mathbf{y}}^{(i)}
    =
    \left(\mathbf{W}_{\mathrm{out}}^{*}\right)^{\top}
    \begin{bmatrix}
        1\\
        \widetilde{\boldsymbol{\xi}}^{(i)}
    \end{bmatrix}.
\end{equation}

where \(\widetilde{\boldsymbol{\xi}}^{(i)}\) is the z-score normalized dynamic-summary vector. Figure~\ref{fig-task2}B illustrates how these quantities are computed from a representative bending sensor response. The mean terms capture the deformation bias induced by the shifted COM, while the standard-deviation terms capture the fluctuation magnitude of the ringdown response. The early and late statistics capture how the response evolves after the shake input, and the late-minus-early drift captures the net change during relaxation. The RMS term captures the overall response magnitude. The calculation of these quantities are detailed in SI Section S3B.

Compared to raw sensing data, the dynamic-summary representation gives the linear readout a much more compact description of the ringdown response. In each experimental trial, the raw sensor reading contains 3000 data points per channel, whereas the dynamic summary contains only 8---calculating the mean and standard deviation is computationally inexpensive and adds only minor overhead. Importantly, readouts based on the dynamic summary produce substantially more accurate predictions of the hidden COM's position (right columns of Figure~\ref{fig-task2}C, D). This improvement indicates that COM information is distributed across the bias, oscillation, decay, and magnitude of the ringdown response, rather than concentrated in a single instant of the signal. This use of physically interpretable response summaries is consistent with reservoir-computing analyses that relate performance to memory, information capacity, and task-relevant observability in dynamical systems~\cite{dambre2012information, love2023spatial, carroll2022optimizing}.

\medskip
To further explain how the dynamic-summary representation supports position-aware COM prediction, we use partial least squares (PLS)~\cite{wold1984collinearity} to visualize the distribution of ringdown data collected at different COM positions. Similar to principal component analysis (PCA)~\cite{jolliffe2016principal}, PLS reduces high-dimensional data to a small number of principal components. The key difference is that PCA determines these components based solely on the variance within the measured data, whereas PLS selects components that maximize the covariance between the data and the target outcome---here, the changing COM position. PLS therefore allows us to examine whether ringdown data from different COM positions form distinct clusters. We apply PLS separately to the vectorized raw ringdown data, \(\operatorname{vec}\left(\mathbf{X}^{(i)}\right)\), and to the dynamic-summary vector, \(\boldsymbol{\xi}^{(i)}\). Denote \(o^{(i)}\) as the true COM position in trial \(i\); the first PLS component for each representation is obtained from

\begin{equation}
\begin{aligned}
\mathbf{v}_{1,\mathrm{raw}}
&=
\underset{\lVert\mathbf{v}\rVert_2=1}{\arg\max}
\operatorname{Cov}_{i}^{2}\!\left(
\mathbf{v}^{\top}\operatorname{vec}\!\left(\mathbf{X}^{(i)}\right),
o^{(i)}
\right),
\\
\mathbf{v}_{1,\mathrm{summary}}
&=
\underset{\lVert\mathbf{v}\rVert_2=1}{\arg\max}
\operatorname{Cov}_{i}^{2}\!\left(
\mathbf{v}^{\top}\boldsymbol{\xi}^{(i)},
o^{(i)}
\right).
\end{aligned}
\label{eq:pls_projection}
\end{equation}

Here, \((i)\) indexes the experimental trial, \(\mathbf{X}^{(i)}\) is the raw time-series matrix defined in Equation~(\ref{eq-reservoir-states}), and \(\boldsymbol{\xi}^{(i)}\) is the dynamic-summary vector defined in Equation~(\ref{eq-summary}). The operator \(\operatorname{vec}(\cdot)\) converts \(\mathbf{X}^{(i)}\) into a column vector. The vector \(\mathbf{v}\) is a unit vector defining a candidate direction in the high-dimensional space, while \(\mathbf{v}_{1,\mathrm{raw}}\) and \(\mathbf{v}_{1,\mathrm{summary}}\) are the directions selected for the first PLS component of the raw and dynamic-summary representations, respectively. The second PLS direction is then obtained in the same manner, after removing the contribution of the first component. Each dot in Figure~\ref{fig-task2}F therefore represents one trial projected onto the first two PLS components. The raw bending-sensor data produce broad, overlapping clusters across different COM positions, whereas the dynamic-summary features produce more compact and better-separated clusters. Thus, by suppressing phase-sensitive variation in the raw data while retaining load-dependent bias, fluctuation strength, transient drift, and response magnitude, the dynamic-summary representation presents COM information to the linear readout in a more usable form.

\medskip
Task II therefore moves beyond hidden-COM orientation classification. The ringdown response obtained from the embedded bending sensors carries cues about the hidden COM's position, while its dynamic-summary representation presents these cues to the linear readout in a more usable form. We next demonstrate how this inferred COM information can be connected to a subsequent robotic action, such as re-grasping.

\subsection{Task III: Use the Inferred COM Position to Inform Subsequent Manipulation Actions}
\label{subsec:level3_com_guided_regrasping}
The final Task III takes a step further, aiming to connect the inferred hidden object physics to manipulation-relevant actions. More specifically, we challenge the soft robotic arm to re-grasp an object according to the inferred COM position to achieve a more balanced lift.  

\begin{figure}[t]
  \centering
  \includegraphics[width=\linewidth]{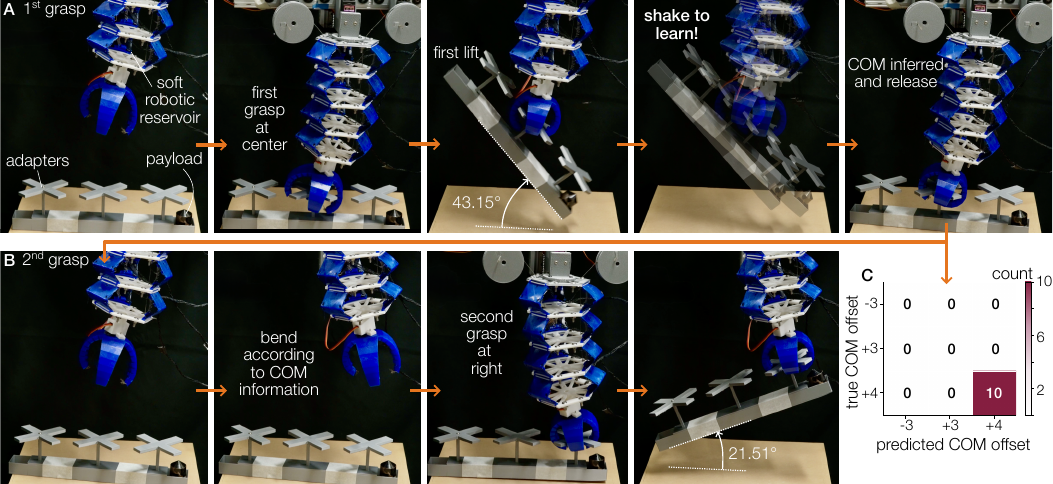}
  \caption{
  Object grasping and re-grasing according inferred center-of-mass positions.  
  (\textbf{A}) During the first grasp, the robot starts from standby, grasps the bar-shaped container at its geometric center, then shakes the bar to infer the center-of-mass position (at $o=+4$ in this case).
  (\textbf{B}) In the second grasp, the robot returns to standby, bends towards the inferred COM position, and then grasp and lift the container again. The bar-container's angle decreased from \(43.15^\circ\) to \(21.51^\circ\).
  (\textbf{C}) Readout accuracy for the COM-offset unseen in training. The \(+4\) condition is repeated ten times to evaluate whether the readout trained on $o=\pm3$ data gives a consistent prediction.
  }
  \label{fig-task3}
\end{figure}

\medskip
Figure~\ref{fig-task3} shows the demonstration sequence. The robot starts from standby, approaches the bar-shaped container (with additional adapters to facilitate grasping), and performs the first grasp and lift at the geometric center. After this first grasp, a brief servo-driven shake input excites the coupled robot-object, and the embedded bending sensors record the ringdown responses. Once the robot infers the object's hidden COM position, it executes a second grasp toward the COM (a 60 g payload mass at $o=4$ on the right in this case).  As a result, the bar tilt angle decreases from \(43.15^\circ\) after the first grasp to \(21.51^\circ\) after the second grasp, showing that the second grasp moves the bar toward a mechanically more balanced configuration.

\medskip
This implementation separates the real-time hidden-COM readout from low-level manipulation motion execution. The approach, first-grasp, shake-to-learn, and second-grasp motions are prescribed motion primitives. The embedded bending sensor reading is acquired and processed---immediately after the ringdown window ends---based on the dynamic summary setup and pre-trained readout. This way, the robot can obtain the inferred COM position with minimal latency, and this inferred COM position determines which motion primitives should be executed next. This framing keeps the actuation sequence simple, without training a full closed-loop control command, while preserving the main purpose of the demonstration. That is, a short dynamic interrogation can reveal hidden physical cues, informing the subsequent robot actions. More details on the implementation sequence, sensing pipeline, and training and testing setting are available in the SI Section~S5 and Video~S1.

\medskip
In addition to grasping and re-grasping, task III serves an additional purpose: to explore the feasibility of extrapolating and predicting hidden COM positions \textit{beyond} the training set. In the previous tasks, the tested COM positions were always the same as, or within the range of, the COM positions used for readout training. For example, in Figure~\ref{fig-task2}E (task II), the readout weights were trained with COM located at $o=\pm3$, and tested with COM at $o=\pm3$ or $o=\pm2$. In Task III, the \(60~\mathrm{g}\) payload mass was placed at $o=\pm3$ for readout training (specifically, 14 training trials collected at \(o=-3\) and 14 trials at \(o=+3\)). After training, the readout weights were held fixed and tested for an unseen, larger-offset COM position at \(o=+4\). This condition was repeated ten times, and the reservoir's output was consistently accurate (Figure~\ref{fig-task3}).

\section{Discussion and Conclusion}
\label{sec:conclusion_outlook}
This work demonstrates a new modality for robotic perception and learning based on interaction-induced dynamics and physical reservoir computing. A servo-driven shaking input excites the soft robotic arm while it grasps an object, causing the object’s hidden center-of-mass (COM) position to be projected, or encoded, into the measurable ringdown response of the coupled robot-object system. These dynamics are captured either through camera tracking or an embedded bending sensor, and a trained linear readout decodes the COM position through a weighted summation of the measured responses. Because the same shaking input is applied in every trial, variations in the observed dynamics arise from the way the grasped object alters the behavior of the coupled system. In this sense, the origami arm serves not only as a sensorized structure, but also as a physical reservoir computer whose intrinsic dynamics transform hidden object physics into a high-dimensional, interpretable response.

\medskip
This study also offers several broader insights and contributions. First, robotic learning through interaction-induced dynamics is largely agnostic to sensor modality. As demonstrated by the Task I results in Figure~\ref{fig-task1}, both camera tracking and embedded bending-sensor measurements yield accurate predictions of COM orientation. The agreement between these sensing pathways indicates that the relevant payload information is encoded in the ringdown dynamics themselves, rather than in any particular measurement technology. The key requirement, therefore, is that the soft robotic body exhibits sufficiently high-dimensional and separable dynamics in response to changes in object physics. Second, we derive a compact dynamic summary representation that preserves deformation bias, fluctuation strength, transient relaxation, and overall response magnitude. As shown by the Task II results in Figure~\ref{fig-task2}, this reduced representation enables quantitatively more accurate prediction of the hidden COM position. Finally, the regrasp experiment in Task III demonstrates that real-time sensing and inference of the COM cue can guide subsequent grasp-related decision-making (Figure~\ref{fig-task3}). To this end, this study presents an easy-to-implement framework in which grasp, shake, and regrasp actions remain open-loop primitives, while the COM offset is computed online from real-time bending-sensor measurements.

\begin{figure}[t!]
  \centering
  \includegraphics[scale=1.0]{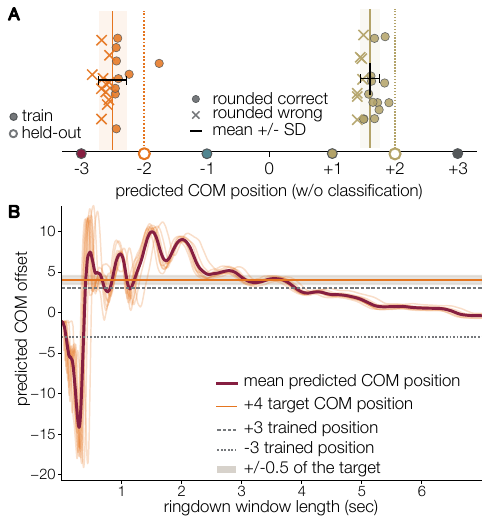}
  \caption{\textbf{Exploratory Studies for Future Investigations}.
    (\textbf{A}) Quantitatively accurate COM-distance prediction. The readout weights are trained on $o=-3,-1,+1,+3$ and tested on held-out $o=-2,+2$ trails. The results reaches $55\%$ exact-cell accuracy, above the seven-cell chance level of $14.3\%$. For target COM positions at $-2$ and $+2$, the predictions were $-2.5\pm0.22$ and $1.6\pm0.16$, respectively. (\textbf{B}) The overall prediction accuracy changes according the the length of the ringdown dynamics recording. 
    }
  \label{fig-future}
\end{figure} 

\medskip
Several directions merit further investigation. First, there remains room to improve the \textit{quantitative} accuracy of COM position prediction. In this study, task success was mainly evaluated using nearest-target classification; however, in more demanding manipulation scenarios, a robot may need a quantitatively accurate estimate of the hidden object’s physical properties. To explore this possibility, we conducted an additional experiment using linear regression for exact-position interpolation rather than nearest-target classification, which is substantially more challenging than the preceding tasks (Figure~\ref{fig-future}A). Specifically, the readout was trained on COM locations at $o=-3,-1,+1$ and $+3$, and then evaluated on held-out trials with COM positions at $o=-2$ and $+2$. Using the dynamic summary representation across 40 trials, the physical reservoir predicted $\hat{o}=-2.5\pm0.22$ and $\hat{o}=1.6\pm0.16$. Although this result is encouraging, further improvements in quantitative generalization may require an enhanced robotic design with richer and more sensitive dynamic responses.

\medskip
Another important topic for future study is the required duration of the recorded ringdown response. In this work, a 3-second recording window produced the most accurate results (Figure~\ref{fig-future}B). Shorter windows were insufficient to capture the relevant reservoir dynamics, whereas longer windows reduced accuracy because the informative dynamics had already decayed. For other robotic platforms and target object properties, the optimal ringdown recording length will likely be an important system parameter that must be determined empirically.

\medskip
Nonetheless, the results of this study suggest a promising new direction for robotic perception, in which hidden physical properties are revealed through interaction and processed through the body dynamics generated by that interaction. By controlling how a robot probes an object and measuring how its compliant body responds afterward, robotic systems may recover latent information about mass distribution, balance, and other mechanics relevant to manipulation. This perspective extends physical reservoir computing beyond body-centered information processing toward object-directed dynamic interrogation. More broadly, it points toward a class of robots whose compliant structures are designed not only for actuation and sensing, but also to render hidden object properties observable through interaction by exploiting their intrinsic physical computing capability.

\medskip
\textbf{Supporting Information} \par
Additional supporting information can be found online in the Supporting Information section.

\medskip
\textbf{Acknowledgements} \par
The authors acknowledge the support from the National Science Foundation (CMMI-2328522 and EFMA-2422340).
\medskip
\bibliographystyle{unsrtnat}
\bibliography{references}

\end{document}